\documentclass[a4paper,twoside]{article}
\usepackage{epsfig}
\usepackage{subcaption}
\usepackage{calc}
\usepackage{amssymb}
\usepackage{amstext}
\usepackage{amsmath}
\usepackage{amsthm}
\usepackage{multicol}
\usepackage{pslatex}
\usepackage{apalike}
\usepackage{algorithm2e}
\usepackage[bottom]{footmisc}
\usepackage{SCITEPRESS}     

\begin{document}

\title{Interpretable Markov-Based Spatiotemporal Risk Surfaces for Missing-Child Search Planning with Reinforcement Learning and LLM-Based Quality Assurance}

\author{\authorname{Joshua Castillo and Ravi Mukkamala\orcidAuthor{0000-0001-6323-9789}}
\affiliation{Old Dominion University, Norfolk, VA, USA}
\email{\{jcast046, rmukkama\}@cs.odu.edu}
}


\abstract{The first 72 hours of a missing-child investigation are critical for successful recovery. However, law enforcement agencies often face fragmented, unstructured data and a lack of dynamic, geospatial predictive tools. Our system, Guardian, provides an end-to-end decision-support for missing-child investigation and early search planning. It converts heterogeneous, unstructured case documents into a schema-aligned spatiotemporal representation, enriches cases with geocoding and transportation context, and provides probabilistic search products spanning 0–72 hours. In this paper, we present an overview of Guardian as well as a detailed description of a three-layered predictive component of the system. The first layer is a Markov chain, a sparse, interpretable model with transitions incorporating road accessibility costs, seclusion preferences, and corridor bias with separate day/night parameterizations. The Markov chain’s output prediction distributions are then transformed into operationally useful search plans by the second layer’s reinforcement learning. Finally, the third layer’s LLM performs post hoc validation of layer 2’s search plans prior to their release. Using a synthetic but realistic case study, we report quantitative outputs across 24/48/72-hour horizons and analyze sensitivity, failure modes, and tradeoffs. Results show that the proposed predictive system with the 3-layered architecture, produces interpretable priors for zone optimization and human review.}

\onecolumn \maketitle \normalsize \setcounter{footnote}{0} \vfill

\section{\uppercase{Introduction}}
\label{sec:introduction}
Missing-search planning is a complex, multidisciplinary process that requires coordinated integration of information, expertise, and resources from multiple parties. The process typically begins with the synthesis of heterogeneous inputs such as last-known-position estimates, environmental and terrain data, weather forecasts, sensor observations, and behavioral or mobility models of the missing subject \cite{USCG2013SAR}. These inputs are provided by various stakeholders, including search-and-rescue coordinators, field teams, experts on the subject, data analysts, and increasingly computational decision-support systems \cite{FrostStone2001SearchTheory}. Effective coordination among these parties enables the assessment of uncertainty, prioritization of search regions, allocation of limited assets, and continuous adaptation of the plan as new evidence emerges \cite{Washburn2018SearchDetection}. Consequently, the effectiveness of missing-search operations depends not only on the accuracy of individual data sources, but also on the structured fusion of multi-party inputs to support timely and informed decision-making under uncertainty.

Missing-child investigations begin with incomplete, rapidly evolving information and limited time to structure a search plan. Traditional search planning is based on human judgment, coarse heuristics, and manual fusion of heterogeneous sources (reports, PDFs, tips, transit information, and maps). In practice, the challenge of early-stage investigation is not only to “predict” a location, but to produce calibrated uncertainty and actionable search products under severe data sparsity \cite{RuizReyes2025MissingReview} \cite{Solaiman2022MissingML} \cite{FBI2014CARP}. In practice, early search planning is bottlenecked not only by limited information but by the time cost of converting fragmented narratives (reports, tips, bulletins, maps) into a shared, geographically grounded plan before the search area expands and resources become diluted. Guardian targets this gap by standardizing and validating case evidence early, then producing interpretable 24/48/72-hour geospatial products (risk surfaces, ranked sectors, and bounded zones) that help coordinators prioritize effort under uncertainty without replacing established investigative procedures.

We have addressed this problem through Guardian, an end-to-end pipeline that converts raw unstructured case documents into a probabilistic search surface over a wide geographical grid and a set of human-interpretable artifacts: ranked sectors, hotspots, and containment rings for 24/48/72-hour horizons. Guardian is organized as a two-stage system (see Figure~\ref{fig:Arch}). In stage 1, the data pre-processing stage (Guardian Parser Pack), the system ingests the provided raw data, normalizes it, validates the data the resulting output, and enriches it with information from other sources. In stage 2, the analysis and evaluation stage (Guardian Core), it goes through a series of steps including data validation, case generation, LLM processing with consensus, formation of clusters and hotspots, forecast and identification of plausible ring-and-likelihood (R\&L) zones, generation of search plans, and  plan evaluation. The resulting outputs are designed to be auditable and consumable by investigators without requiring the internals of the model. 

In this paper,  we provide an overview of the Guardian system, with emphasis on the predictive aspects of the system from stage 2. The predictive component is designed as a three-layer architecture. The first layer is a  Markov mobility forecasting component, which connects the extracted case context to geospatial priors and produces sequential horizon forecasts with explicit probability conservation within the given geographical boundaries, and survival-style temporal decay. 

The second layer contains a reinforcement learning model that converts belief maps provided by the Markov model into a compact set of actionable search zones. Finally, the LLM in the third layer validates the search plans through a quality assurance process before their release.

A running synthetic case study (GRD-2025-001541) is used throughout to illustrate how outputs evolve across horizons and how corridor vs. seclusion effects shape the forecast.

The paper is organized as follows. In section 2, we summarize related work. Section 3 describes the Guardian's overall system architecture, including a description of the two stages. In section 4, we provide the details of the Markov mobility forecasting model including the underlying mathematical foundations. Section 5 summarizes layer 2 of the prediction system, reinforcement learning. In section 6, layer 3 of the prediction system, the LLM quality assurance layer is discussed. Section 7 provides details of the implementation of the predictive system. Section 8 summarizes the results of a specific case study, known as GRD-2025-001541. Finally, Section 9 summarizes our contribution and provides plans for future work.

\section{\uppercase{Related Work}}

Data-driven approaches to missing-person search have expanded from descriptive statistics to predictive modeling, emphasizing the need for a spatial output that is aware of uncertainty \cite{RuizReyes2025MissingReview}. Data fusion for missing-person problems has also been framed as an end-to-end pipeline challenge with heterogeneous sources, variable reliability, and operational constraints \cite{Solaiman2022MissingML}.

Markov chains have been widely used to model movement over transportation networks and spatial grids due to interpretability and tractable uncertainty propagation (Besenczi et al., 2021). Recent Search and Rescue (SAR)-specific mobility algorithms integrate GIS context to improve plausibility \cite{Papic2024MobilitySAR}. Similar research efforts in this direction include inverse optimal control and activity forecasting that combine environmental features with trajectory inference \cite{Kitani2012ActivityForecasting}, and agent-based models that incorporate behavioral profiles and terrain constraints to produce probabilistic movement distributions \cite{Hashimoto2022LostPersonABM}.

Unsupervised spatial clustering supports the discovery of recurring regions of interest and is frequently used for crime and incident hotspot analysis \cite{Longley2015GIS}. Kernel- and density-based priors provide an empirical base rate over geography that can regularize sparse case-specific forecasts \cite{Longley2015GIS}.

Reinforcement learning (RL) has been applied to time-critical search settings (e.g., drone-enabled SAR) to allocate limited resources over dynamic belief maps \cite{Ewers2024DRLSAR}. Reward shaping methods support stable policy learning when objectives combine coverage efficiency and early capture \cite{Kwon2025AdaptiveReward}.

Converting unstructured narrative reports into structured fields is fundamental in police and investigative analytics \cite{Chau2002PoliceNarratives}. Large Language Models (LLMs) can act as constrained annotators and weak labelers when guided by schema validation, rather than as unconstrained end-to-end extractors \cite{Chen2024LLMAnnotator}. Weak supervision frameworks enable the creation of scalable labels via heuristic functions and validation \cite{Ratner2017Snorkel}.

Missing-person systems operate on sensitive data. Hence, responsible design requires safeguards for privacy, mitigation of misuse, and a decision-support framework \cite{ICRC2025MissingPeopleTech}.

In this paper, we integrate Markov models with reinforcement learning and LLMs to provide validated and feasible search plans for missing-person location planning.

\section{\uppercase{System Architecture}}
\begin{figure*}[h]
  \centering
  \includegraphics[width=\linewidth]{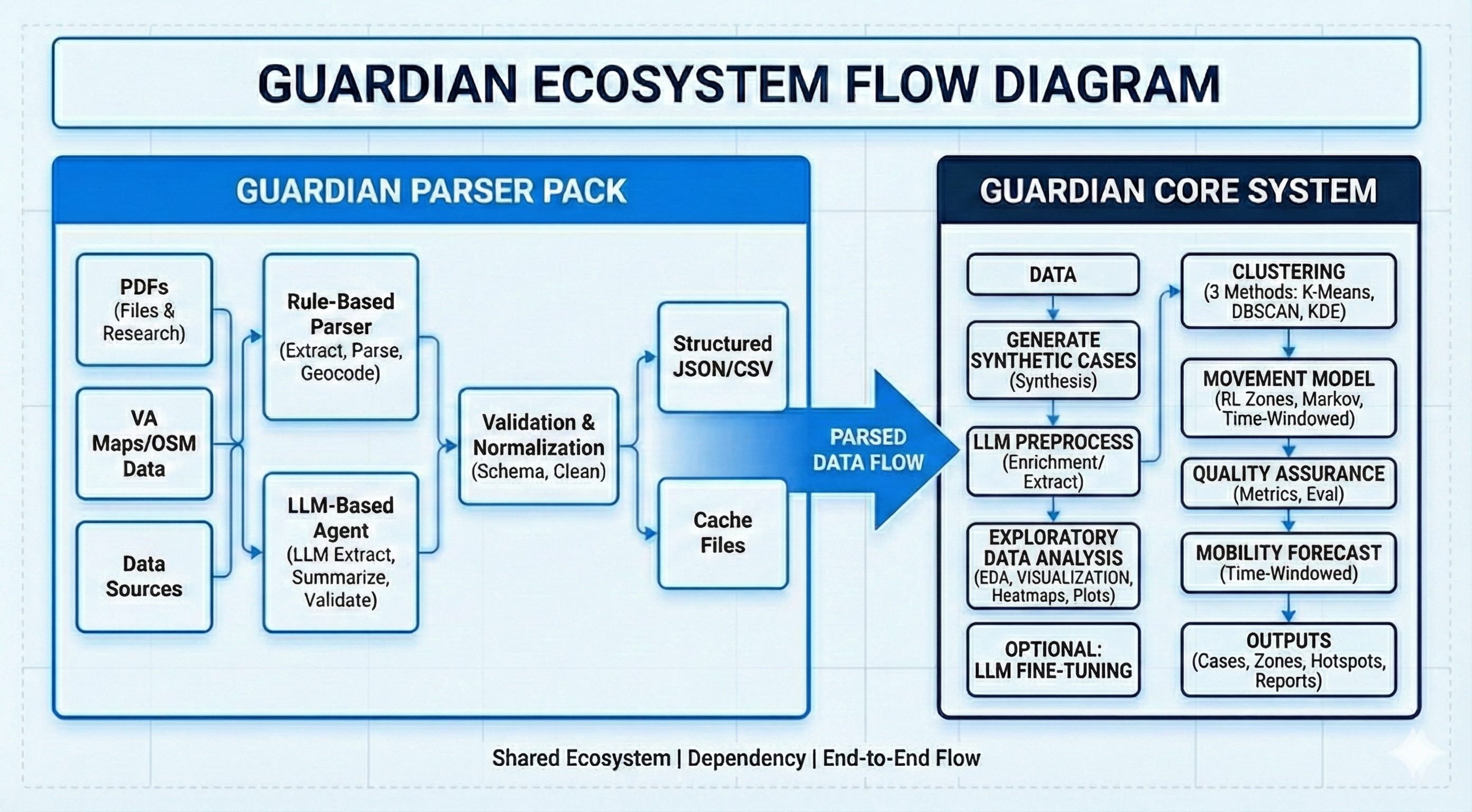}
  \caption{Guardian System Architecture with two distinct but interconnected systems}
  \label{fig:Arch}
\end{figure*}

As shown in Figure~\ref{fig:Arch}, Guardian is organized as a modular ecosystem with the Guardian Parser Pack and the Guardian Core System, designed to operate independently, but interlock through a shared data contract and a unidirectional auditable data flow.

The Parser Pack is the data preparation and standardization subsystem. Its primary responsibility is to convert heterogeneous and unstructured missing-person PDF documents into machine-readable records that conform to schema and are suitable for modeling and evaluation. Inputs include investigative PDFs (e.g., NamUs/NCMEC/FBI/Charley Project/VSP-style reports) and geospatial resources (e.g., map layers, gazetteers, road/transit context).

The Parser Pack runs a hybrid extraction pipeline. It performs multi-engine text extraction (e.g., PDF text engines with OCR fallback for scanned documents), and then applies source-aware parsing (rule-based templates when formats are stable, and LLM-assisted extraction when narratives are variable). This reflects a general best practice in investigative analytics with deterministic extraction providing reliability on known structures, while learned language components provide coverage on noisy narratives \cite{Chen2024LLMAnnotator}.

The Parser Pack enforces a single shared schema as a contract between ingestion and downstream modeling. Records are normalized (canonical field names, standardized timestamps, consistent categorical values), enriched (geocoding, county/state identifiers, proximity to transportation features) and then validated against schema constraints. When the LLM path is used, validation failures can trigger targeted repair loops (re-prompting only the failing fields) to reduce hallucinated or malformed outputs and increase traceability \cite{Chen2024LLMAnnotator}. It outputs structured datasets (e.g., JSONL/CSV) and maintains cache artifacts (notably, geocoding caches and intermediate extraction outputs) to support reproducibility and performance. In short, the Parser Pack can be run as a standalone “PDF-to-structured-records” system, but in Guardian it functions as the authoritative upstream producer of clean, schema-valid case data.

 The Core System is the analysis, prediction, and decision-support subsystem. It consumes schema-valid case records and transforms them into probabilistic search products over 24/48/72-hour horizons, including risk surfaces, ranked sectors, hotspot lists, and containment rings. The Core System can operate in two modes: (i) using Parser-Pack-produced records from PDFs, or (ii) using synthetic, schema-conformant cases generated to provide ground truth for controlled experimentation without exposing sensitive personal data \cite{Sun2023SyntheticHealth}. Synthetic case generation emphasizes geographic realism (within specified geographical boundaries and road/transit adjacency) and narrative realism so that downstream extraction and forecasting are evaluated under conditions that resemble operational noise.

Within the Core System, multiple modeling stages run in a coordinated pipeline. First, an LLM preprocessing layer produces constrained extraction, summarization, and weak labels (e.g., type of movement, risk level) to convert narrative cues into structured signals while retaining deterministic checks and consensus/ quality control behaviors \cite{Chen2024LLMAnnotator}. Next, clustering and hotspot detection build empirical spatial priors using complementary methods (K-Means, DBSCAN, and KDE) so that the system can represent both coarse regional modes and density-connected clusters while smoothing uncertainty into usable probability fields \cite{Ester1996DBSCAN} \cite{Longley2015GIS}. 

These priors feed the mobility engine, which forecasts location distributions via a sparse, interpretable Markov propagation over a specified gridded geographical space. Transitions incorporate geospatial features (e.g., transportation accessibility, corridor bias, seclusion preference), possibly parameterized differently for day/night dynamics. to capture short-horizon non-stationarity while preserving tractable uncertainty propagation \cite{Besenczi2021TrafficMarkov} \cite{Papic2024MobilitySAR}. Temporal calibration is handled through survival-style decay/half-life weighting, so uncertainty appropriately widens with time rather than remaining overconfident \cite{RuizRizzo2022OlderAdults} \cite{RuizReyes2025MissingReview}.

The resulting belief maps are then converted into actionable plans by a reinforcement-learning (RL) zone optimizer, which frames search planning as a sequential allocation problem under resource constraints, balancing early capture value against area/overlap penalties \cite{Ewers2024DRLSAR} \cite{Kwon2025AdaptiveReward}. A dedicated quality assurance layer evaluates outputs using operationally meaningful metrics (e.g., Geo-hit@K, time-to-first-hit, cluster stability, predictive consistency) and supports optional LLM-based zone plausibility checks to reduce nonsensical recommendations. 

The Core System ultimately produces investigator-facing artifacts (ranked sectors, rings, hotspot summaries, and reports), all explicitly positioned as decision support, not necessarily as an autonomous authority source, consistent with responsible AI governance guidance for high-stakes humanitarian and public-safety contexts \cite{Floridi2019AIEthics} \cite{ICRC2025MissingPeopleTech}.

The two systems, the Parser Pack and the Core system, interact through a strict parsed data flow, where Parser Pack outputs flow into the Core System as inputs. They share (i) a common case schema (the contract enabling compatibility and validation), (ii) common geospatial dependencies (geographical boundary masks, road/transit layers used both in enrichment and in mobility transitions), and (iii) caching/logging conventions that support reproducibility. This separation of concerns—clean ingestion vs. probabilistic modeling—keeps the architecture extensible, where the Parser Pack can evolve with new document formats, while the Core System can evolve with new forecasting/optimization methods without breaking the end-to-end pipeline.

In the following section, we describe Markov mobility forecasting model in detail.

\section{\uppercase{Markov Mobility Forecasting}}

Layer 1 of Guardian’s prediction system uses Markov modeling to forecast the likely location distribution of a missing person in future horizons (24 h, 48 h, 72 h). It has five inputs: (i) a seed distribution around the last known point, (ii) a previous hotspot, (iii) Markov propagation over a geographical grid with transportation-informed transitions, (iv) survival-style temporal decay, and (v) boundary masking. In this section, we describe the details of this Markov mobility forecasting  system. The notation used in this section is summarized in Table 1.
\begin{figure*}[h]
  \centering
  \includegraphics[width=\linewidth]{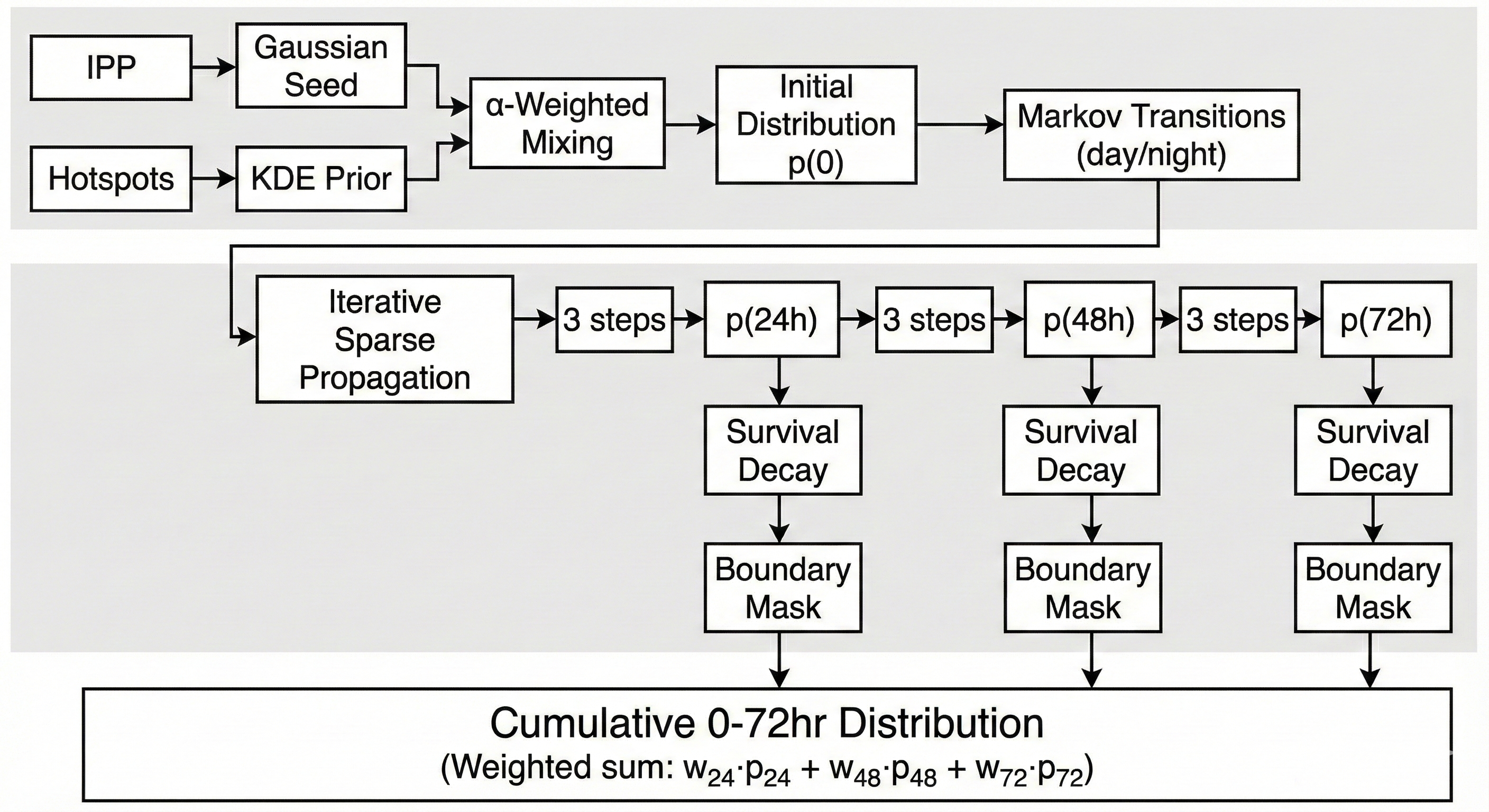}
  \caption{Markov mobility forecasting pipeline (seed, prior, transitions, decay).}
  \label{fig:Markov}
\end{figure*}

\begin{table}[htbp]

\begin{tabular}{p{2cm} p{4.5cm}}
\hline
\textbf{Symbol} & \textbf{Meaning} \\
\hline

$N$ & Number of grid cells \\

$\mathcal{N}(i)$ & Neighbor set of cell $i$ \\

$X_t$ & Latent location at time $t$ \\

${p}_t$ & Probability vector at time $t$ \\

$p_t(i)$ & Probability of cell $i$ at time $t$ \\

$\Delta^{N-1}$ & Probability simplex \\


IPP & Initial Planning Point \\
$T_{1/2}$&Half-life parameter\\


$\sigma$ & Gaussian spread parameter \\

${h}$ & Hotspot prior \\

KDE & Kernel Density Estimation \\

$\alpha$ & Mixing weight \\

${p}_0$ & Initial distribution \\

${P}$ & Normalized transition matrix \\

$\tilde{P}$&Unnormalized transition matrix\\

${P}^T$ & Transposed transition matrix \\

$K$ & Number of propagation steps \\

$w_{ij}$ & Unnormalized transition weight \\

${p}_{24},{p}_{48},{p}_{72}$ & Horizon forecasts \\


$\lambda$ & Survival decay weight \\

${m}$ & Boundary mask \\

$\odot$ & Element-wise multiplication \\

${p}_{\text{cum}}$ & Cumulative distribution \\

$\beta_h$ & Horizon weight \\
$\gamma_H$ & Earlier horizon weight \\

$i, j$ & Grid cell indices in propagation \\

$d(i,j)$ & Distance between cells \\

$c_\text{road}({j})$ & Road accessibility cost \\

$s_{\text{sec}}(j)$ & Seclusion score \\

$c_{\text{corr}}(j)$ & Corridor score \\

$\beta_d, \beta_r, \beta_s, \beta_c$ & Feature weights \\

$j'$ & Dummy neighbor index \\

${p}_h$ & Horizon forecast \\
$\varsigma$&Case information\\

\hline
\end{tabular}
\centering
\caption{Notation used in Markov Mobility Forecasting System}
\label{tab:symbol_definitions_markov}
\end{table}

The Markov Mobility Forecasting Pipeline estimates where a missing person may be over time by gradually spreading and refining uncertainty in a controlled, interpretable way. A high-level view of the pipeline is shown in Figure~\ref{fig:Markov}. (A detailed state transition diagram of the Markov chain is omitted due to limited space). 

The forecasting process begins with two simple probability sources: a Gaussian seed centered on the Initial Planning Point (IPP), which reflects uncertainty around the last known location by assigning higher likelihood for nearby locations and lower likelihood for farther away locations, and a historical hotspot prior built using kernel density estimation over past hotspot patterns identified by clustering methods such as K-Means and DBSCAN. 

These two inputs are mixed through a mixture $\alpha$-weighted to form a normalized starting distribution   that balances what is known about the current case with broader historical tendencies. This distribution is then advanced forward in discrete steps using Markov transition matrices, which describe how probability can move between locations while conserving total probability and allowing different movement behavior during day and night periods. Repeated propagation produces forecast maps at 24, 48, and 72 hours (referred to as 24 h, 48 h, and 72 h, respectively), each adjusted using survival-style decay to reflect growing uncertainty over time and boundary masking to restrict probability to valid geographic areas.

The resulting horizon-specific forecasts are finally combined into a cumulative 0–72 hour distribution. The probabilistic distributions are then transformed into  practical search products such as ranked search sectors, distance-based containment rings, and prioritized hotspot regions, using layer 2's reinforcement learning and layer 3's quality assurance by LLMs. The details of Layers 2 and 3 are presented in Sections 5 and 6, respectively.

Figure 2 also makes explicit the two-stage structure of the predictor: (i) initialization and (ii) time-forward propagation with per-horizon post-processing. The top row shows how the model constructs the starting belief $p(0)$ by combining case-local evidence (IPP → Gaussian seed) with a population-level tendency (hotspots → KDE prior), then blending them via image.png-weighted mixing to produce a single normalized distribution. The middle row highlights that forecasting is performed by iterative sparse propagation (implemented as a small fixed number of Markov steps per horizon) using day/night transition matrices, which preserves probability mass while shifting it along plausible local moves defined by the transition structure. 

The bottom row emphasizes that each horizon map is regularized and constrained before downstream use: survival decay dampens long-horizon confidence to reflect increasing uncertainty, and the boundary mask enforces geographic validity (e.g., Virginia containment) so probability does not accumulate in inadmissible cells. Finally, the figure clarifies how the system produces both horizon-specific outputs image.png , image.png and a cumulative 0–72h risk surface via a weighted sum $w_{24}p_{24}+w_{48}p_{48}+w_{72}p_{72}$ which is the distribution used to derive sectors, containment rings, and RL/QA inputs.

\subsection{Problem formulation}
We discretize the search domain into $N$ grid cells within the given geographical area. We define a latent location state $X_t\in \lbrace 1,\ldots,N\rbrace$ at discrete time step $t$. The goal is to compute a probability vector $p_t \in \Delta^{N-1}$, where $p_t(j)=Pr(X_t=j| \varsigma)$ given case information  $\varsigma$ (last seen, movement profile, context, priors).

Guardian uses a first-order Markov assumption for computational and interpretability reasons with $P_{ij}=Pr(X_{t+1}=j | X_t=i,\varsigma)$, where $P$ is a row-stochastic transition matrix. 

The propagation $p_t$ is defined as: $p_{t+1}=P^T p_t \Rightarrow p_{t+k}= (P^T)^k p_t$.
This formulation yields efficient multi-step forecasting and interpretable transition influences \cite{Besenczi2021TrafficMarkov} \cite{Papic2024MobilitySAR}.

\subsection{Transition matrix construction}

A central design requirement of the Markov chain's transition matrix $P$ is to: (i) incorporate transportation plausibility, (ii) remain sparse, and (iii) be interpretable. The Guardian constructs $P$ on a fixed grid using the adjacency of the nearest neighbors (KNN) $K$. Each cell $i$ connects to a small neighbor set $N(i)$ of size $k$, producing a sparse matrix with $O(Nk)$ non-zeros.

\subsubsection{Feature layers and costs}

Each cell $j$ has features derived from the specified geospatial context:
\begin{itemize}
\item Road accessibility cost $c_{road}(j)$, where the lower cost indicates an easier movement.
\item Seclusion score $s_{sec}(j)$, where a higher score indicates more concealment in terrain/land use.
\item Corridor score $c_{corr}(j)$, where the higher score indicates proximity to major corridors (e.g., interstates).    
\end{itemize}
	
These features are constructed from road/ traffic layers and environmental proxies and are bounded and normalized for stability \cite{Longley2015GIS}.

\subsubsection{Energy-based transition weights}
We employ energy-based transition weights in the Markov chain as a way to turn costs, preferences, and constraints into probabilistic transitions in a principled, interpretable manner. These are especially useful when we want transitions to reflect how easy or plausible a move is, rather than learning opaque probabilities. Here, we define them as follows.


\subsubsection{Row-stochastic normalization and sparsity}
Row-stochasticity ensures probability conservation:

$$P_{ij}=\begin{cases}
    \frac{\tilde{P_{ij}}} 
    {
    \sum_{j'\in{N(i)}} 
    \tilde{P_{ij'}}
    } & j\in N(i)\\
    0 &otherwise
\end{cases}$$
where $\tilde{P_{ij}}$ is the unnormalized and $P_{ij}$ is the normalized energy-based transition weights that encode relative movement preference between neighboring grid cells $i$ and $j$. Because each row has only $k$ candidates, computing and storing $P$ is $O(Nk)$, enabling fast horizon propagation.

\subsubsection{Day/night transition matrices}
The system uses separate day/night matrices $P^{day}$ and $P^{night}$ with different parameterizations of $\beta$. This supports differing mobility patterns (e.g., reduced travel and increased concealment at night). During forecasting, the system selects the appropriate matrix per time step using local time derived from the last-seen timestamp and horizon step. This explicitly introduces short-horizon non-stationarity while retaining a Markov form.

\subsubsection{Gaussian Seeding}
Forecasting starts from an initial belief $p_0$. It is composed of a spatial seed placed at the last-seen point (IPP) and diffused as an isotropic Gaussian over grid cells:
$$p_{seed}(j)\propto exp(-\frac{d(IPP,j)^2} {2\sigma^2})$$
where $\sigma$ is selected conservatively based on movement profile (e.g., “on foot” vs. vehicle cues) and reporting delay and $d(i,j)$ is the geographic distance between cell centers of grids $i$ and $j$ used to penalize implausible long-range transitions.  Here $\propto$ implies that the probability assigned to the location $j$ follows the shape of this exponential curve, but is not yet normalized to sum up to 1.

Guardian builds a hotspot prior  $p_{prior}$ from historical/synthetic cases and multi-method clustering. This prior captures base-rate geography, where disappearances historically concentrate, independent of the current case.
The initial distribution is a convex mixture:
$$p_0=(1-\alpha_{prior})p_{seed}+\alpha_{prior} p_{prior}$$
followed by renormalization. $\alpha_{prior}$ prevents overconfidence when the last-seen location is noisy or when reporting delays are long, while still preserving case specificity.

\subsection{Survival-style temporal decay}
Empirically, uncertainty should widen over time, but probability in implausible regions should remain down-weighted. Guardian introduces a profile-dependent half-life decay applied at each horizon, capturing the decreasing probability of remaining within the modeled mobility envelope as time elapses \cite{RuizRizzo2022OlderAdults} \cite{RuizReyes2025MissingReview}. Operationally, the system applies a multiplicative decay factor
$w(t)=2^{-t/T_{1/2}}$, where  $T_{1/2}$ is a half-life parameter

To preserve a normalized spatial distribution for decision-support, decay is implemented as a calibration weight (affecting cumulative horizon blending and zone thresholds) while the displayed grid distribution is renormalized after masking. This maintains interpretability (a proper distribution over space) while allowing time-based down-weighting in multi-horizon fusion.

\subsection{Boundary masking and geographic constraints}
The specified geographical boundaries are enforced through a mask $m(j)\in\lbrace0,1\rbrace$. After each propagation step or at each horizon output, probabilities outside the boundary are zeroed, and the distribution is renormalized:

$$p\leftarrow \frac {m \bigodot p} {\Sigma_jm(j)p(j)}$$ where $\odot$ denotes element-wise multiplication

This prevents probability leakage to unreachable regions, improving interpretability and operational alignment.

\subsection{Sequential horizon propagation and cumulative forecast}
The system computes sequential horizons $H\in\lbrace24,48,72\rbrace$ hours. Let $k_H$ be the number of Markov steps to represent $H$. Forecasts are produced sequentially:
\begin{itemize}
	\item Compute $p_{24}=(P^T)^{k_{24}}*p_0$
	\item Use $p_{24}$ as the start for 48h: $p_{48}= (P^T)^{k_{24}}*p_{24}$
	\item Similarly, for 72h: $p_{72}= (P^T)^{k_{24}}* p_{48}$
\end{itemize}

A cumulative forecast for 0–72h can be constructed as a weighted blend:

$$p_{(0-72)}\propto\sum_{H\in\lbrace24,48,72\rbrace}\gamma_H *w(H)p_H$$

with weights $\gamma_H$ chosen to emphasize earlier horizons in time-critical search settings.


\section{Reinforcement Learning (RL)}
Layer 2 of Guardian’s prediction system employs reinforcement learning to translate probabilistic belief maps generated by the Markov mobility model  into a compact and actionable set of search recommendations. While the Markov layer estimates where a missing child is likely to be over time, the RL layer addresses the operational question of how to search, framing search-zone selection and prioritization as a sequential decision-making problem under resource and spatial constraints.

At each decision step, the RL agent observes the current belief state produced by the Markov model, represented either as a full spatial probability surface or as aggregated sector-level probability masses, along with the active time horizon and predefined resource limits. Based on this observation, the agent selects a small set of search actions, instantiated as candidate zones such as circular regions centered on high-probability cells or hotspots, sector-wide allocations, or ring-band sweeps derived from cumulative probability contours. The action space is intentionally constrained to ensure interpretability and operational feasibility, reflecting real-world limits on personnel and search capacity.

The reward function combines three complementary objectives. First, an early capture reward incentivizes covering a high-probability mass as early as possible in the 0 to 72 hour window, reflecting the time-critical nature of missing-child investigations (Ewers et al., 2024). Second, a coverage efficiency term penalizes redundant overlap between zones and excessively large search areas, encouraging parsimonious plans that maximize information gain per unit effort. Third, plausibility shaping incorporates soft constraints aligned with the Markov forecast, such as consistency with corridor proximity, seclusion preference, and geographic validity (e.g., remaining within state boundaries). 

The RL layer operates over three fixed temporal windows—--0–24 hours (weight 1.0), 24–48 hours (weight 0.7), and 48–72 hours (weight 0.5)—and produces ranked outputs for each window, including prioritized sectors, candidate search zones, and containment rings. The confinement rings are calculated directly from the cumulative belief distribution and reported in standard quantiles (50\%, 75\%, and 90\%) to provide an intuitive summary of spatial uncertainty. Importantly, while the Markov layer provides the probabilistic movement model, the RL layer functions purely as a decision layer: it consumes belief maps and produces search actions without modifying the underlying transition dynamics. This decoupling preserves modularity and interpretability, allowing the assumptions and outputs of each layer to be independently audited.



Adaptive reward shaping is used to stabilize learning across case types.
The RL layer outputs ranked sectors, candidate zones, and containment rings for communication and planning. Here, the containment rings are derived from the cumulative distribution and reported in standard quantiles (50/75/90\%).



\section{LLM-based Quality Assurance}
Layer 3 of Guardian’s prediction system introduces a Large Language Model–based Quality Assurance (LLM QA) component that performs post hoc validation of the search plans generated by the reinforcement learning layer. Although Markov and RL components produce search zones that are probabilistically optimal under the assumptions modeled, such as mobility priors, accessibility to transport, corridor bias, seclusion preference, and survival-style decay—statistical optimality alone does not guarantee investigative plausibility. In operational contexts, purely quantitative models may occasionally propose zones that are mathematically defensible but inconsistent with narrative constraints, behavioral cues, or real-world investigative expectations that are difficult to encode explicitly in the state space.

To address this gap, Guardian incorporates an LLM QA layer that evaluates each candidate search zone in context. The QA module receives a structured case summary, the inferred movement profile, the relevant time window, and the geometric properties of each zone and produces a bounded plausibility score along with a brief rationale in natural-language. This process serves as a semantic and contextual sanity check, identifying zones that may conflict with narrative details (e.g., implausible travel modes or contradictions with known constraints) even if they carry non-trivial probability mass.

The LLM QA system is implemented using lightweight, instruction-tuned models—specifically Qwen-2.5-3B-Instruct and LLaMA-3.2-3B-Instruct—selected for their efficiency, determinism, and suitability for audit-sensitive environments. These models operate under schema-constrained prompting to ensure consistent outputs and avoid unconstrained generation. Crucially, the plausibility scores produced by the LLM do not alter the Markov transition matrices or the RL reward formulation. Instead, they are applied as a final reweighting of zone priorities, allowing Guardian to retain probabilistic rigor while improving semantic coherence, transparency, and human interpretability of the resulting search plans.

In combination, the three-layer architecture positions the Markov model as the predictive core, reinforcement learning as the operational decision layer, and LLM-based QA as a validation and calibration layer. This separation of concerns enables Guardian to produce search recommendations that are not only statistically grounded, but also aligned with investigative reasoning and suitable for human-in-the-loop decision support.



\section{\uppercase{Implementation}}
In this section, we describe our experimental setup to test the proposed system and the implementation details.

We chose the Commonwealth of Virginia for our case study. The state is discretized into a masked grid; transitions are defined via KNN adjacency to maintain sparsity. We have ingested the past reported cases from available PDF documents via Parser Pack. These have been enriched with roads/transit/corridor/seclusion layers and passed through clustering priors and Markov forecasting. LLM components operate deterministically with schema constraints and optional validation-driven repair loops  \cite{Chen2024LLMAnnotator}.

The performance of the system is measured in terms of the following metrics.
\begin{itemize}
    \item Sector mass (\%): aggregated probability mass in operational sectors.
    \item Containment radii: radii that capture a probability of 50/75/90\% around the IPP (useful to communicate uncertainty).
    \item Hotspot concentration: cumulative probability mass captured by top-K hotspots (K=50) to quantify spread over time.
    \item Illustrative evaluation metrics (pilot): Geo-hit@K and area-searched-until-hit (ASUH) for ablations \cite{Lyu2025IGUIDE}, clearly labeled as illustrative trends when not measured directly in the case study.
\end{itemize}

To anchor ‘best practice’ comparisons, we report results against two operational baselines already produced by the system: (i) Markov+RL zone priorities derived purely from probabilistic mass and reward shaping (baseline ranking), and (ii) the same zones re-ranked by the LLM plausibility layer (QA-adjusted ranking), which functions as an auditable post-hoc validation rather than a new predictor. Evaluation is framed in search-centric terms familiar to both SAR and IR audiences—coverage and prioritization—using Geo-hit@K and median best distance (does the top-K set cover the ground-truth point within zone radii), plus optional effort proxies such as area-searched-until-hit and time-to-first-hit (how early in the 0–72h windows a plan would likely succeed). The system has been implemented in Python 3.x, using NumPy/SciPy for numerical and spatial kernels, GeoPandas/Shapely for GIS geometry operations, and scikit-learn for clustering/KDE. The end-to-end run for a case follows the following principles: (i) structured case loading, (ii) geospatial layer loading, (iii) Markov forecast generation, (iv) RL zone selection, and (v) LLM-based zone quality assessment (QA) and reweighting.

In the current implementation, we have limited our prediction state space to a fixed Virginia-wide grid. All feature layers are aligned by grid index. 
Primary inputs include grid cell center, derived by sampling a regular grid over a Virginia bounding box and applying a Virginia boundary mask to retain in-domain cells; accessibility cost for each cell (lower = easier access), derived from the Virginia road network (e.g., OSM road segments) via bounded Dijkstra shortest-path/distance-to-network computations, then normalized; encoding proximity to major corridors (e.g., interstates), derived from distance-to-corridor features (corridor polylines) and a monotone decay mapping; encoding concealment preference (higher $\Rightarrow$ more secluded), derived from a weighted and normalized composite of environmental/context layers (e.g., population density inverse and land-use/POI proxies); and 
prior inputs based on previous cases including hotspots.

In terms of the synthetic case (referred to as GRD-2025-001541), the initial planning point (IPP) (last-seen lon/lat/time), behavioral tags (e.g., on-foot), and follow-up sightings were used for evaluation/zone scoring.

The quantitative forecast artifacts are read from several files: search\_plan.json (sector masses, hotspots, containment rings, horizon forecasts, IPP),
 zones\_review (LLM QA plausibility and priority changes) and zone-qa-metrics (case-level QA summary statistics).
 
The system emits two primary machine-readable products, a Search Plan JSON and a Zones JSONL stream.

The search plan output contains the following features: ipp---last-seen coordinates (initial planning point); grid\_xy---list of coordinates;
p---cumulative 0–72h probability over grid cells (aligned to grid\_xy indices);
per-horizon distributions (e.g., "24", "48", "72"), each aligned to grid\_xy;
sectors\_ranked---probability mass aggregated by sector (cumulative and per horizon),
and rings---containment radii for specified quantiles (e.g., 50/75/90\%).

\begin{figure*}[!h]
  \centering
  \includegraphics[width=\linewidth]{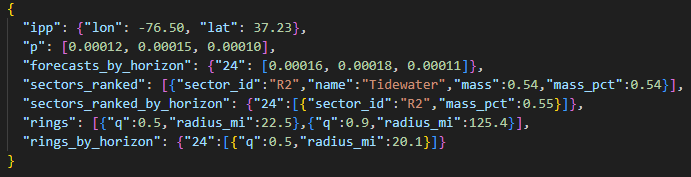}
  \caption{Example Conference Search Plan Output}
  \label{fig:SearchPlan}
\end{figure*}

 An example of the system search plan output is shown in Figure~\ref{fig:SearchPlan}. The output variables in the plan can be interpreted as follows. 
\begin{itemize}
    \item Index alignment: $p[i]$ is the probability assigned to the location grid\_xy[i].
    \item Sector mass: mass\_pct is the fraction of total probability within that sector (higher $\Rightarrow$ higher regional priority).
    \item Rings: q=0.5 gives the radius around the IPP that contains 50 percent of the probability mass (uncertainty summary).
    \item Horizons: forecasts\_by\_horizon["24"] is the 24h belief map; later horizons should broaden (controlled by transitions + masking + decay calibration).
\end{itemize}
	
We now look at the Zone outputs. It consists of case\_id; zones---zones grouped by time window (e.g., ``0-24", ``24-48", ``48-72"), each zone with center\_lon, center\_lat, radius\_miles, and a priority score; reward/zone\_scores (for RL scoring traces); QA\_outputs---plausibility, rationale, and original\_priority vs. new\_priority.

\begin{figure*}[h]
  \centering
  \includegraphics[width=\linewidth]{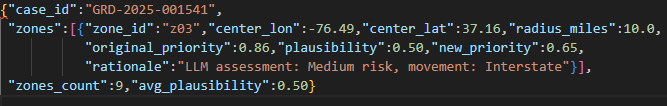}
  \caption{Example Zone QA Reweighted output}
  \label{fig:ZoneQA}
\end{figure*}

As mentioned in section 6, the LLM QA system is implemented using lightweight, instruction-tuned models—specifically Qwen-2.5-3B-Instruct and LLaMA-3.2-3B-Instruct—selected for their efficiency, determinism, and suitability for audit-sensitive environments. These models operate under schema-constrained prompting to ensure consistent outputs and avoid unconstrained generation. An example of QA-reweighted output is shown in Figure ~\ref{fig:ZoneQA}. Here, original\_priority is derived from the Markov forecast (probability mass at/near the center of the zone); new\_priority is the LLM-QA adjusted priority used for the final ranked zone list; and plausibility is a semantic calibration scalar from narrative analysis (bounded [0,1]) applied consistently across zones in that case.

\section{\uppercase{Results}}
This section reports quantitative results for the required running case study GRD-2025-001541 and clearly labeled illustrative ablation trends from pilot synthetic evaluations. GRD-2025-001541 is a synthetic but operationally realistic Virginia missing-child case generated by the Guardian schema-validated case generator to enable repeatable evaluation while protecting real victims’ identities; the generator constructs cases using real geographic and transportation data and behavioral patterns derived from historical missing-child investigations and child-abductor/trafficker criminal psychology case studies extracted through the Guardian Parser Pack from investigative reports and research sources. This case is representative because it includes a last-seen anchor (IPP), realistic behavioral movement profiles (e.g., residential-local/on-foot dynamics informed by prior case patterns), and optional follow-up sightings, allowing the full Guardian pipeline—from structured evidence extraction through Markov mobility forecasting, zone generation, and QA validation—to be evaluated across the 24/48/72-hour search horizons under conditions that approximate real investigative complexity.

We evaluated our system using GRD-2025-001541, a synthetic but realistic missing-child case involving a 15-year-old female last seen in York, Virginia, at 03:58 AM, characterized as residential-local movement, on foot, within a custodial context. Forecasts were generated at 24h, 48h, and 72h horizons using the full Guardian pipeline. Figures 5-7 show 24/48/72-hour maps. We now illustrate these maps in more detail by referring to Figure 5.

Figure 5 visualizes the 24-hour forecast as a probability surface over Virginia, where darker (redder) grid cells indicate a higher likelihood and lighter cells indicate a lower likelihood of presence. The red star marks the Initial Planning Point (IPP), anchoring the forecast to the last-seen location, while black markers denote historical hotspots that inform the prior. The probability field is not uniform or radial; instead, it concentrates along transportation-connected regions and known activity clusters, reflecting the incorporation of road accessibility, corridor bias, and seclusion preferences of the Markov mobility model. 

Rectangular sector overlays summarize how probability mass aggregates operationally, showing that the Tidewater sector captures the majority of near-term likelihood, with Northern Virginia emerging as a strong secondary region due to corridor connectivity rather than simple distance diffusion. The dashed blue rings represent containment radii (e.g., 50\% and 90\%), providing an intuitive measure of uncertainty: at 24 hours, half of the modeled probability lies within a relatively compact radius around the IPP, while the outer ring illustrates plausible but lower-probability tail risk. 

Taken together, the map shows that early-horizon uncertainty remains localized but structured—probability spreads in a controlled, interpretable way along plausible movement pathways rather than dispersing uniformly—supporting prioritized, regionally focused search planning rather than exhaustive statewide coverage.

Across all forecast horizons, the probability mass remained strongly concentrated in the Tidewater region, consistently retaining more than half of the total forecast probability ($>$50\%). This reflects the combination of a localized movement profile, early-morning disappearance timing, and dense historical hotspot priors in southeastern Virginia.
Northern Virginia emerged as the dominant secondary region, accounting for approximately one quarter to one third of the total probability mass (~24–30\%), driven primarily by the connectivity of the corridor rather than direct local diffusion. All remaining regions individually contributed comparatively small probability shares.

As the forecast horizon increased from 24 to 72 hours, the spatial uncertainty expanded in a structured manner. Probability distributions diffused outward from the initial point while preserving the corridor-aligned structure rather than collapsing to uniformity. This behavior reflects the short-horizon non-stationarity of the Markov process combined with transportation constraints.

The containment ring analysis further illustrated this diffusion. The 50\% probability containment radius expanded from roughly 20 miles at 24 h to the mid-20-mile range by 72 h, indicating a gradual but controlled spatial spread. Despite this expansion, the majority of probability mass remained geographically localized, supporting prioritization of early-stage regional search allocation.

Observed follow-up sightings embedded in the synthetic case narrative fell predominantly within high-probability regions identified by the forecast, particularly within the Tidewater area at earlier horizons. Later sightings aligned with regions adjacent to the moderate-probability corridor, consistent with the diffusion dynamics of the model.

\begin{figure}[h]
  \centering
  \includegraphics[width=\linewidth]{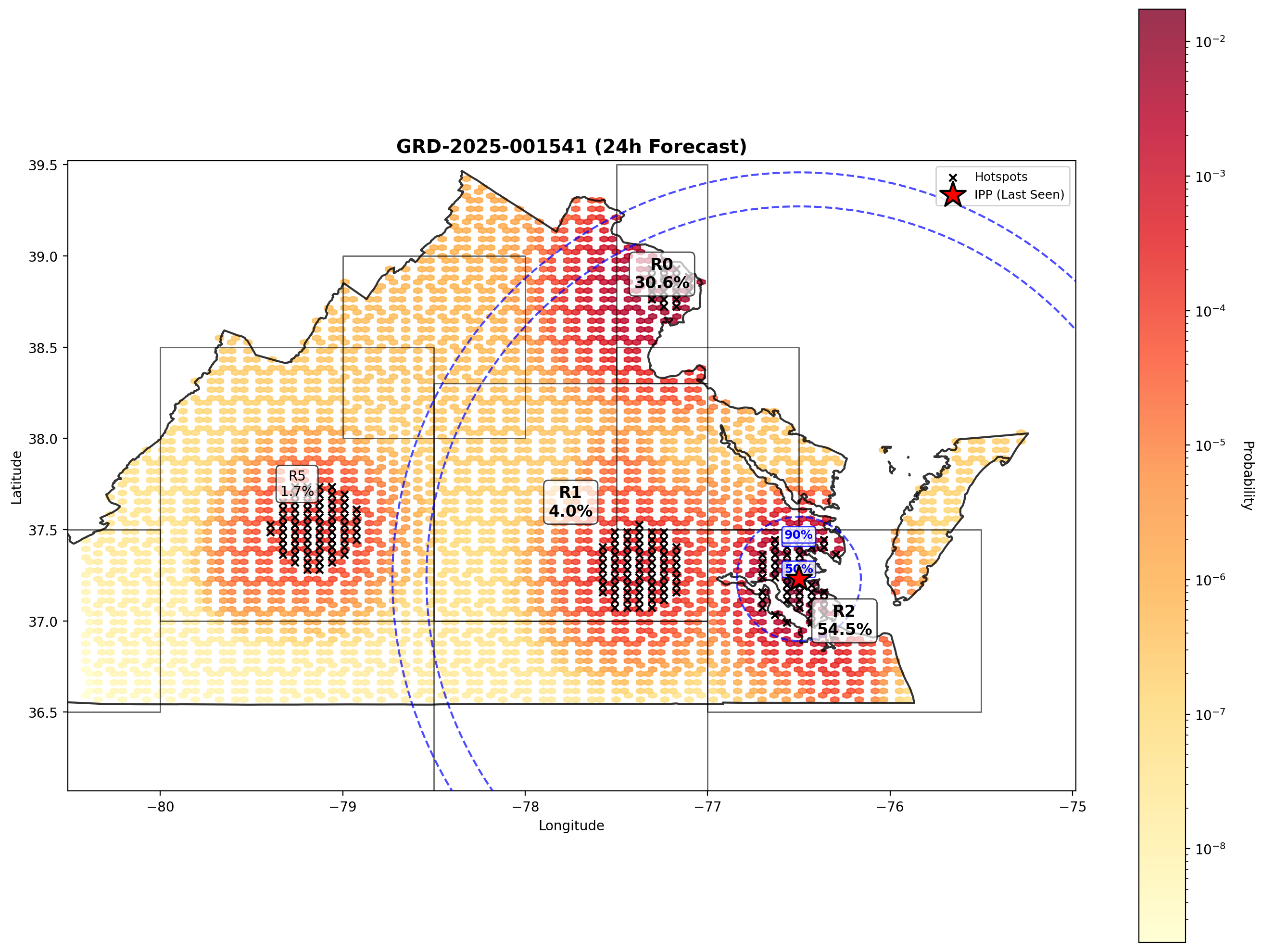}
  \caption{24-hour Search Forecast}
  \label{fig:24h}
\end{figure}

\begin{figure}[h]
  \centering
  \includegraphics[width=\linewidth]{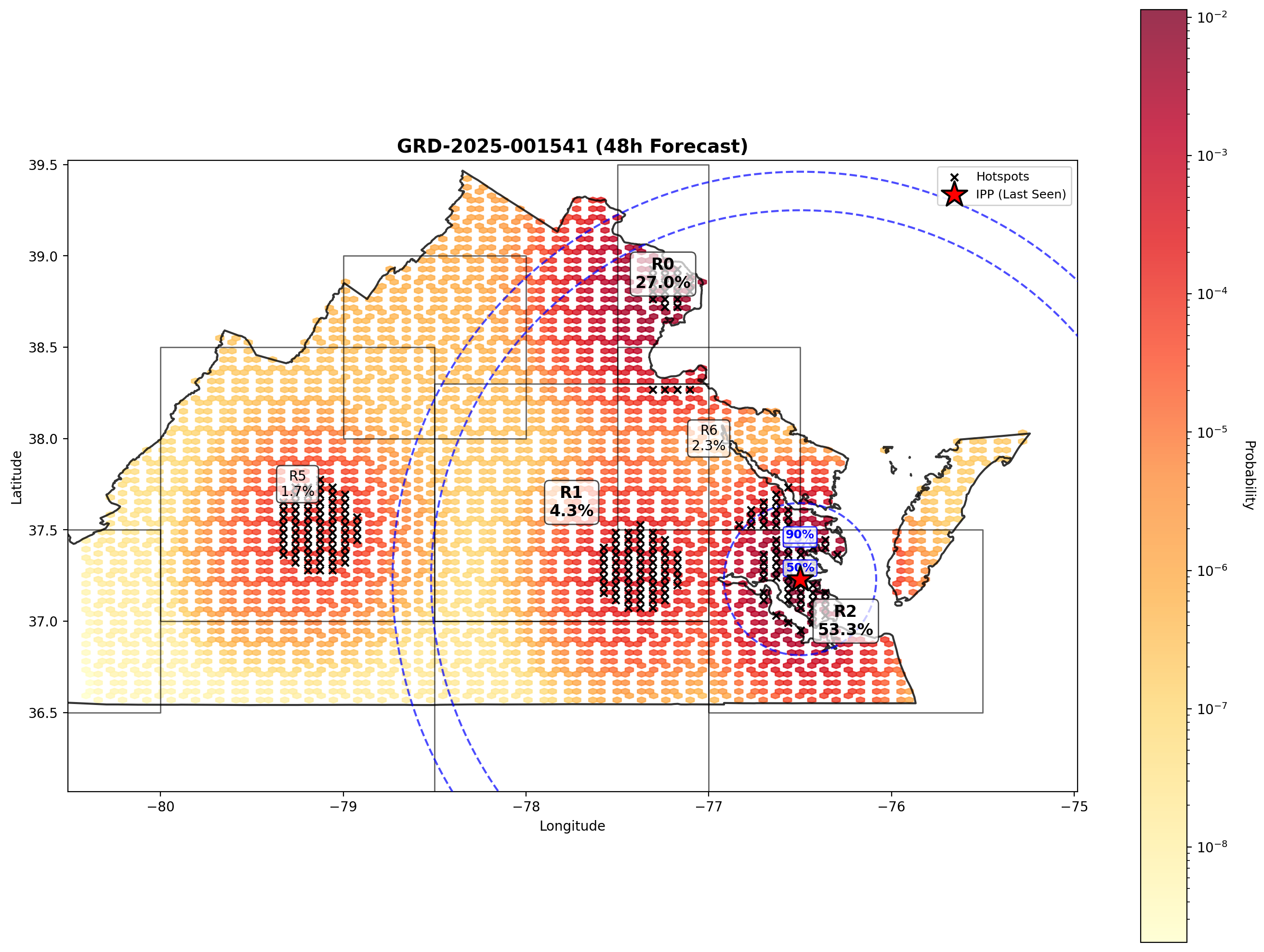}
  \caption{48-hour Search Forecast}
  \label{fig:48h}
\end{figure}

\begin{figure}[h]
  \centering
  \includegraphics[width=\linewidth]{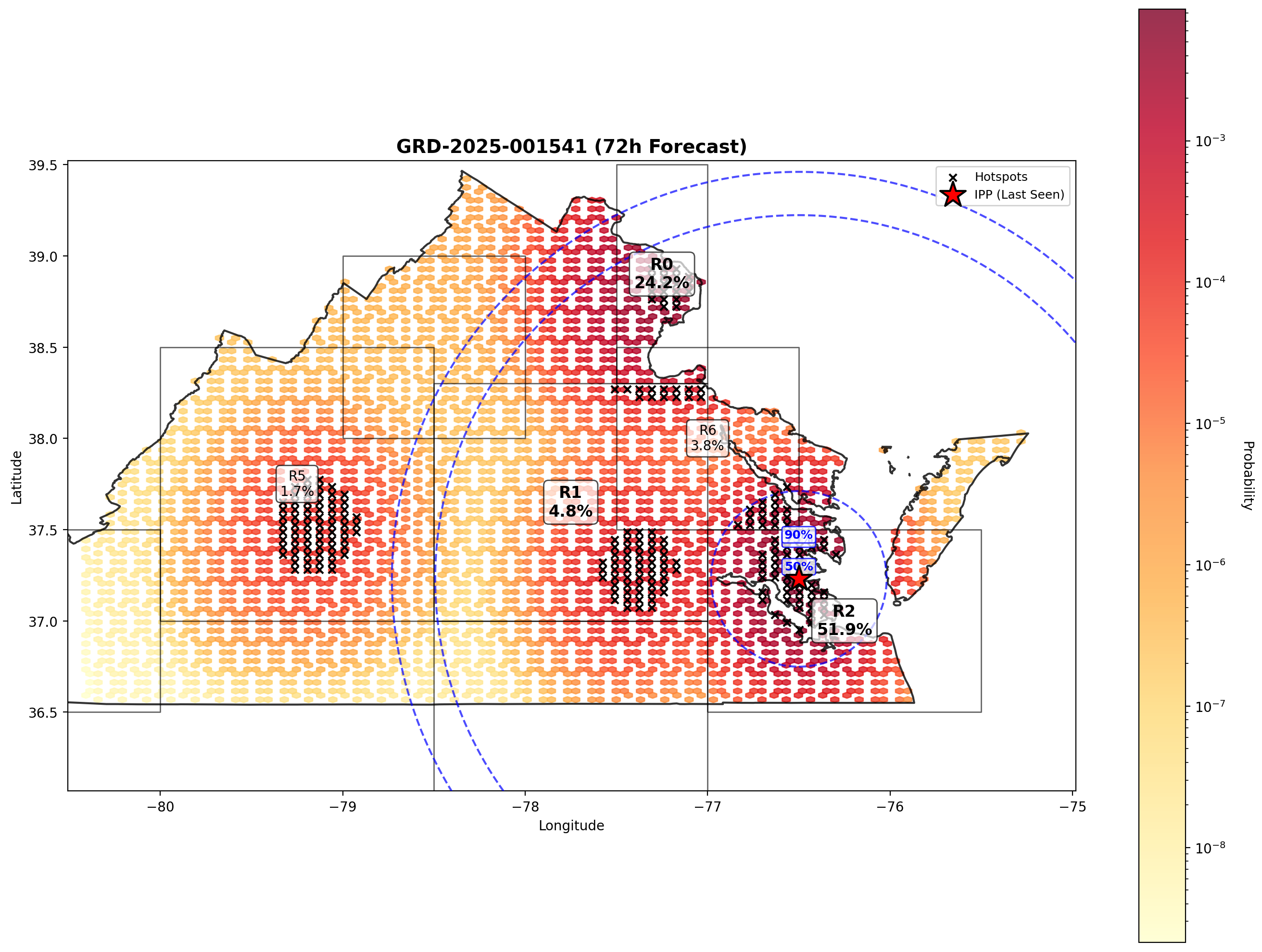}
  \caption{72-hour Search Forecast}
  \label{fig:72h}
\end{figure}

From our experiments, we make several observations. 
\begin{itemize}
\item Impact of the horizon. Horizon increases compound uncertainty, resulting in more Markov steps that diffuse mass. Night-/day switching changes transition preferences and survival-style weighting reduces late-horizon dominance in cumulative planning. Consequently, the concentration of the hotspot decreases over time, reflecting a shift from sharp local peaks to broader plausible regions.
\item Corridor vs. seclusion behavior. Guardian’s transition weights explicitly trade corridor bias against seclusion reward. In urban/suburban Tidewater contexts, corridor proximity can increase plausibility for rapid displacement even under on-foot priors, while seclusion rewards support hiding-location hypotheses. The sustained dominance of Tidewater in the case suggests a local region connected to the corridor that remains plausible across the horizons, while the mass of Northern Virginia indicates diffusion along the connectivity pathways and base-rate priors.
\item Failure modes. We have seen several causes for the predictions of the system to fail. These include
incorrectly mislocalized last-seen coordinates, mis-specified profile (on foot vs. vehicle), over-reliance on biased historical priors \cite{RuizReyes2025MissingReview} \cite{ICRC2025MissingPeopleTech}, and 
	LLM hallucination in extraction of data \cite{Chen2024LLMAnnotator} \cite{Li2025LLMEvaluation}.
\item Sensitivity. The most sensitive components of the system are found to be (i) $\alpha_{prior}$ (controls prior vs. case anchoring), (ii) corridor/seclusion weights $\beta_c, \beta_s$, and (iii) day/night switching schedules. Sensitivity analysis should be standard in deployment and outputs should remain advisory rather than prescriptive
\end{itemize}

\section{\uppercase{Conclusions}}
\label{sec:conclusion}
Project Guardian demonstrates a holistic pipeline from raw PDF ingestion to search-zone products for missing-child investigations in Virginia, centered 
on a sparse, interpretable Markov mobility forecasting engine that integrates transportation costs, corridor bias, seclusion rewards, day/night dynamics, 
survival-style calibration, and boundary masking. The required case study GRD-2025-001541 illustrates stable sector dominance (Tidewater ~52–55\% across 
horizons), decreasing hotspot concentration with time, and containment radii that communicate core vs. tail uncertainty.

Limitations follow directly from modeling assumptions: the Markov layer is memoryless and grid-resolved, outcomes depend on geocoding quality and priors, and forecasts can shift under key parameters (e.g., $\alpha_{prior}$, corridor/seclusion weights, day/night schedule), so sensitivity checks are necessary and outputs should be treated as advisory decision support rather than prescriptive automation. Operationally, the pipeline is designed to be re-run when new evidence arrives (e.g., a credible sighting updates the seed/constraints and regenerates forecasts and zones), and while this paper focuses on missing-child profiles, the framework generalizes to other populations (e.g., elderly) by calibrating movement and temporal profiles rather than changing the overall architecture—an approach that emphasizes modularity, interpretability, and auditability as transferable design principles.

Future work includes a systematic calibration on real-world retrospective cases under privacy safeguards, learning $\beta$ parameters from data via inverse modeling while preserving interpretability, integrating higher-order Markov or semi-Markov dynamics when evidence supports non-memoryless movement, stronger uncertainty quantification and domain shift handling between synthetic and real cases, and a rigorous evaluation of multi-LLM consensus efficacy under controlled adversarial extraction conditions.





\bibliographystyle{apalike}
{\small
\bibliography{example}}

@manual{USCG2013SAR,
author = {USGC},
  title        = {U.S. National Search and Rescue Supplement to the International Aeronautical and Maritime Search and Rescue Manual},
  organization = {U.S. Coast Guard},
  address      = {Washington, DC, USA},
  year         = {2013}
}

@techreport{FrostStone2001SearchTheory,
  author      = {Frost, John R. and Stone, Lawrence D.},
  title       = {Review of Search Theory: Advances and Applications to Search and Rescue Decision Support},
  institution = {U.S. Coast Guard Research and Development Center},
  address     = {Groton, CT, USA},
  number      = {CG-D-15-01},
  year        = {2001}
}

@book{Washburn2018SearchDetection,
  author    = {Washburn, Alan R.},
  title     = {Search and Detection},
  edition   = {5},
  publisher = {Springer},
  address   = {Cham, Switzerland},
  year      = {2018}
}

@article{Besenczi2021TrafficMarkov,
  author  = {Besenczi, R. and B{\'a}tfai, N. and Jeszenszky, P. and Major, R. and Monori, F. and Isp{\'a}ny, M.},
  title   = {Large-Scale Simulation of Traffic Flow Using Markov Model},
  journal = {PLOS ONE},
  volume  = {16},
  number  = {2},
  pages   = {e0246062},
  year    = {2021}
}

@article{Chau2002PoliceNarratives,
  author  = {Chau, M. and Xu, J. J. and Chen, H.},
  title   = {Extracting Meaningful Entities from Police Narrative Reports},
  journal = {Journal of the American Society for Information Science and Technology},
  volume  = {53},
  number  = {11},
  pages   = {984--995},
  year    = {2002}
}

@inproceedings{Chen2024LLMAnnotator,
  author    = {Chen, R. and Qin, C. and Jiang, W. and Choi, D.},
  title     = {Is a Large Language Model a Good Annotator for Event Extraction?},
  booktitle = {Proceedings of the Thirty-Eighth AAAI Conference on Artificial Intelligence (AAAI-24)},
  pages     = {17772--17780},
  year      = {2024}
}

@inproceedings{Ester1996DBSCAN,
  author    = {Ester, M. and Kriegel, H.-P. and Sander, J. and Xu, X.},
  title     = {A Density-Based Algorithm for Discovering Clusters in Large Spatial Databases with Noise},
  booktitle = {Proceedings of the Second International Conference on Knowledge Discovery and Data Mining (KDD-96)},
  pages     = {226--231},
  year      = {1996}
}

@misc{Ewers2024DRLSAR,
  author       = {Ewers, J.-H. and Anderson, D. and Thomson, D.},
  title        = {Deep Reinforcement Learning for Time-Critical Wilderness Search and Rescue Using Drones},
  howpublished = {arXiv preprint arXiv:2405.12800},
  year         = {2024}
}

@book{FBI2014CARP,
  author    = {{Federal Bureau of Investigation}},
  title     = {Child Abduction Response Plan: An Investigative Guide},
  edition   = {3},
  publisher = {U.S. Department of Justice},
  year      = {2014}
}

@article{Floridi2019AIEthics,
  author  = {Floridi, L. and Cowls, J.},
  title   = {A Unified Framework of Five Principles for AI in Society},
  journal = {Harvard Data Science Review},
  volume  = {1},
  number  = {1},
  year    = {2019}
}

@article{Hashimoto2022LostPersonABM,
  author  = {Hashimoto, A. and Heintzman, L. and Koester, R. and Abaid, N.},
  title   = {An Agent-Based Model Reveals Lost Person Behavior Based on Data from Wilderness Search and Rescue},
  journal = {Scientific Reports},
  volume  = {12},
  pages   = {5873},
  year    = {2022}
}

@techreport{ICRC2025MissingPeopleTech,
  author      = {{International Committee of the Red Cross}},
  title       = {Balancing Risks and Opportunities: New Technologies and the Search for Missing People},
  institution = {ICRC},
  year        = {2025}
}

@inproceedings{Kitani2012ActivityForecasting,
  author    = {Kitani, K. M. and Ziebart, B. D. and Bagnell, J. A. and Hebert, M.},
  title     = {Activity Forecasting},
  booktitle = {European Conference on Computer Vision (ECCV)},
  pages     = {201--214},
  publisher = {Springer},
  year      = {2012}
}

@misc{Kwon2025AdaptiveReward,
  author       = {Kwon, M. and ElSayed-Aly, I. and Feng, L.},
  title        = {Adaptive Reward Design for Reinforcement Learning},
  howpublished = {arXiv preprint arXiv:2412.10917},
  year         = {2025}
}

@misc{Li2025LLMEvaluation,
  author       = {Li, T. and Qin, Y. and Sheng, O. R. L.},
  title        = {A Multi-Task Evaluation of LLMs' Processing of Academic Text Input},
  howpublished = {arXiv preprint arXiv:2508.11779},
  year         = {2025}
}

@book{Longley2015GIS,
  author    = {Longley, P. A. and Goodchild, M. F. and Maguire, D. J. and Rhind, D. W.},
  title     = {Geographic Information Science and Systems},
  edition   = {4},
  publisher = {Wiley},
  year      = {2015}
}

@inproceedings{Lyu2025IGUIDE,
  author    = {Lyu, F.},
  title     = {Evaluating the Evaluation Matrices: Integrating Spatial Assessment in Geospatial AI Model Training and Evaluation},
  booktitle = {I-GUIDE Forum},
  year      = {2025}
}

@article{Papic2024MobilitySAR,
  author  = {Papi{\'c}, V. and {\v{S}}ari{\'c} Gudelj, A. and Milan, A. and Mili{\v{c}}evi{\'c}, M.},
  title   = {Person Mobility Algorithm and Geographic Information System for Search and Rescue Missions Planning},
  journal = {Remote Sensing},
  volume  = {16},
  number  = {4},
  pages   = {670},
  year    = {2024}
}

@article{Ratner2017Snorkel,
  author  = {Ratner, A. and Bach, S. H. and Ehrenberg, H. and Fries, J. and Wu, S. and R{\'e}, C.},
  title   = {Snorkel: Rapid Training Data Creation with Weak Supervision},
  journal = {Proceedings of the VLDB Endowment},
  volume  = {11},
  number  = {3},
  pages   = {269--282},
  year    = {2017}
}

@article{RuizReyes2025MissingReview,
  author  = {Ruiz Reyes, J. and Congram, D. and Sirbu, R. A. and Floridi, L.},
  title   = {Where Are They? A Review of Statistical Techniques and Data Analysis to Support the Search for Missing Persons},
  journal = {Forensic Science International},
  volume  = {376},
  pages   = {112582},
  year    = {2025}
}

@article{RuizRizzo2022OlderAdults,
  author  = {Ruiz-Rizzo, A. L. and Archila-Mel{\'e}ndez, M. E. and Gonz{\'a}lez Veloza, J. J. F.},
  title   = {Predicting the Probability of Finding Missing Older Adults Based on Machine Learning},
  journal = {Journal of Computational Social Science},
  volume  = {5},
  number  = {2},
  pages   = {1303--1321},
  year    = {2022}
}

@article{Solaiman2022MissingML,
  author  = {Solaiman, K. M. A. and Sun, T. and Nesen, A. and Bhargava, B. and Stonebraker, M.},
  title   = {Applying Machine Learning and Data Fusion to the ``Missing Person'' Problem},
  journal = {IEEE Computer},
  volume  = {55},
  number  = {6},
  pages   = {40--55},
  year    = {2022}
}

@article{Sun2023SyntheticHealth,
  author  = {Sun, C. and van Soest, J. and Dumontier, M.},
  title   = {Generating Synthetic Personal Health Data Using Conditional Generative Adversarial Networks Combining with Differential Privacy},
  journal = {Journal of Biomedical Informatics},
  volume  = {143},
  pages   = {104404},
  year    = {2023}
}



\end{document}